\documentclass{article}

\usepackage{arxiv}

\usepackage[utf8]{inputenc} % allow utf-8 input
\usepackage[T1]{fontenc}    % use 8-bit T1 fonts
\usepackage{hyperref}       % hyperlinks
\usepackage{url}            % simple URL typesetting
\usepackage{booktabs}       % professional-quality tables
\usepackage{amsfonts}       % blackboard math symbols
\usepackage{microtype}      % microtypography
\usepackage{lipsum}		% Can be removed after putting your text content
\usepackage{graphicx}
\usepackage{amsmath,amssymb,amsfonts}
\usepackage{algorithmic}
\usepackage{graphicx}
\usepackage{textcomp}
\usepackage{latexsym}
\usepackage{multirow}
\usepackage{array}
\usepackage{here}
\usepackage{algorithm}
\usepackage{amssymb}

\newcommand{\ub}[1]{\underline{\textbf{#1}}}
\newcommand{\ux}[1]{\underline{#1}}

\title{Octave Mix: Data augmentation using frequency decomposition for activity recognition}

\author{ \href{https://orcid.org/0000-0002-0768-1406}{\includegraphics[scale=0.06]{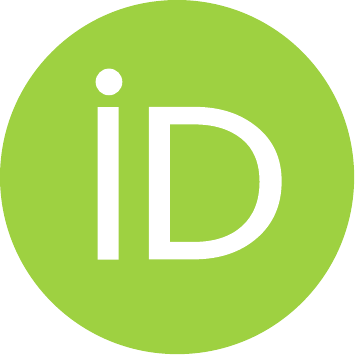}\hspace{1mm}Tatsuhito Hasegawa}\thanks{Research gate url: https://researchmap.jp/t-hasegawa?lang=en} \\
  Graduate School of Engineering,\\
  University of Fukui\\
	Fukui, Japan \\
	\texttt{t-hase@u-fukui.ac.jp} \\
}

\renewcommand{\shorttitle}{\textit{arXiv} Template}

\hypersetup{
pdftitle={Octave Mix: Data augmentation using frequency decomposition for activity recognition},
pdfsubject={},
pdfauthor={Tatsuhito Hasegawa},
pdfkeywords={Data augmentation, Activity recognition, Ensemble deep learning},
}

\begin{document}
\maketitle

\begin{abstract}
	In the research field of activity recognition, although it is difficult to collect a large amount of measured sensor data, there has not been much discussion about data augmentation (DA). In this study, I propose Octave Mix as a new synthetic-style DA method for sensor-based activity recognition. Octave Mix is a simple DA method that combines two types of waveforms by intersecting low and high frequency waveforms using frequency decomposition. In addition, I propose a DA ensemble model and its training algorithm to acquire robustness to the original sensor data while remaining a wide variety of feature representation. I conducted experiments to evaluate the effectiveness of my proposed method using four different benchmark datasets of sensing-based activity recognition. As a result, my proposed method achieved the best estimation accuracy. Furthermore, I found that ensembling two DA strategies: Octave Mix with rotation and mixup with rotation, make it possible to achieve higher accuracy.
\end{abstract}

% keywords can be removed

\section{Introduction}
Task of recognizing human activity by measuring movement from sensors carried by people is called Human Activity Recognition (HAR). By being able to recognize people's activity, lifelogging and the provision of services based on acvivity can be available. Collecting a large amount of activity data of many people make it possible to exploit for marketing and traffic jam mitigation.

\begin{figure*}[tb]
  \centering
  \includegraphics[width=17.0cm]{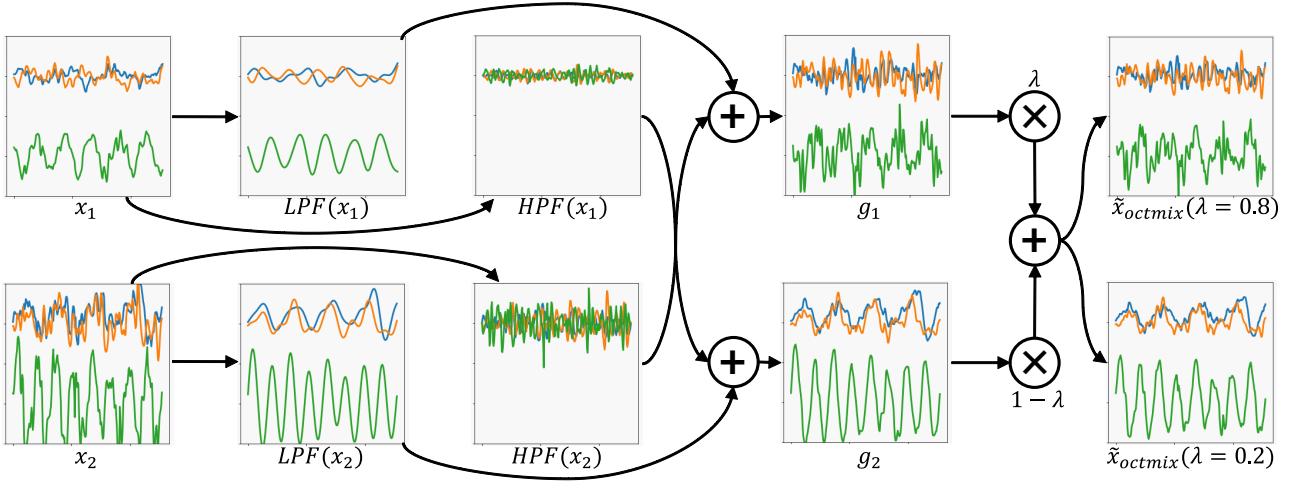}
  \caption{Process of data augmentation by Octave Mix. }
  \label{fig:octmix}
\end{figure*}

HAR is often implemented by machine learning \cite{Lala2012}, and HAR based on deep learning has been actively studied in recent years \cite{jwang2019}. Due to the high expressive ability of deep learning models, a large amount of training data is required to acquire a generic model that avoids overfitting. To address this challenge, data augmentation (DA) \cite{shorten2019} is generally used to expand amout of data in the research field of image recognition. On the other hand, in the field of context-awareness using sensors, it is not easy to expand the labeled data. For example, Inoue et al. \cite{inoue2016} collected a dataset of sensor-based nursing activity recognition. In their study, 22 nurses wearing sensors were performed their nursing nursing tasks, and another nurse as an observer manually recorded their activities. 41 different activity class labels were annotated to the sensor data measured over two weeks. Therefore, it takes a lot of time and effort to collect the training data for HAR, and improving DA methods is desirable.

In this study, I propose Octave Mix (Fig. \ref{fig:octmix}) as a new DA method for HAR using sensor data. In addition, I propose an ensemble model and training algorithm combining the existing DA methods to improve the estimation accuracy. There are some kinds of DA, such as simple geometric transformation style DA for single data and synthetic style for multiple data. Octave Mix is a synthetic-style DA method. After applying a Low Pass Filter (LPF) and a High Pass Filter (HPF) to two sensor data, the two are combined by intersecting low and high frequency data and calculated weighted sum of both combined data. Finally, Octave Mix is used in conjunction with existing DA methods based on geometric transformations. Based on the above, the main contribution of this study is the proposal of the following three methods.

\begin{itemize}
    \item {\bf Octave Mix:} I propose a new synthetic-style DA method, Octave Mix, for sensor data based on frequency decomposition.
    \item {\bf DA ensemble model:} I propose a deep learning model that ensembles multiple DA methods, and investigate the optimal combination of DA for HAR.
    \item {\bf DA revisited for fixed feature extractor (DAR-FFE):} I propose a training algorithm for the proposed ensemble model that applies pre-training to enhance the effectiveness of DA.
\end{itemize}

In addition to the above proposals, my experimental results showed following beneficial findings.

\begin{itemize}
    \item It was found that synthetic-style DA methods (mixup \cite{Zhang2018}, and RICAP \cite{Takahashi2018}) work well for HAR, and that Octave Mix outperforms them.
    \item The best accuracy was achieved by combining two DA policies: Octave Mix with Rotation which is one of the DA methods based on geometric transformations, mixup with Rotation.
\end{itemize}

\section{Related works}
\subsection{Sensor-based human activity recognition}
Most of the research in HAR is based on extracting human-designed features (HCF: Hand-Crafted Features) from sensor values, and classifying the activity using machine learning algorithms such as SVM and Random Forest \cite{Kourosh1998, Bao2004, Ravi2005, Roggen2010, Kwapisz2011, Zappi2012, Reiss2012a, Anguita2013, Bulling2014, Banos2015, Xu2015, Shoaib2016, Daniela2017, Robert2019}. 

With the wide spread of deep learning, end-to-end learning methods, including feature extractor as a trainable network, have been developed. Many studies have been published on HAR studies using deep learning. Many of these studies have simply applied convolutional neural networks (CNNs) to sensor-based HAR tasks\cite{Zeng2014, Chen2015, Yang2015, Ha2015, Gjoreski2016, Hannink2017} The model architecture is based on several convolution-pooling layers followed by a fully-connected layer. A method combining multiple sensors \cite{Yang2018}, methods inserting recurrent layer after several convolution-pooling layers \cite{Francisco2016, Li2018}, and more advanced methods introducing Inception, Residual, and Attention modules \cite{Zhao2018, Dong2019, Long2019, Xu2019, kwang2019} have been studied. On the other hand, these studies focus on the discussion of the optimal model architecture for sensor-based HAR tasks, and do not discuss DA.

Some studies adopted DAs for sensor-based HAR using deep learning \cite{Um2017, Eyobu2018, Kalouris2019, Faridee2019, Rashid2019}. However, their adpted DAs are only folllowings:

\begin{itemize}
    \item {\bf Rotation:} rotate x, y, z axes.
    \item {\bf Permutation:} swap sections in time series.
    \item {\bf Scaling:} scale waveform to amplitude direction.
    \item {\bf Time-warping:} scale waveform to time direction.
    \item {\bf Magnitude-warping:} multiply smooth curve.
    \item {\bf Jittering:} add noise.
    \item {\bf Cropping:} mask a section.
\end{itemize}

Therefore, the discussion of DAs in HAR is limited to the simple geometric transformations. Effectiveness of synthetic-style DAs have not been discussed.

\subsection{Synthetic-style DA}
According to Shorten et al. \cite{shorten2019}, DAs of image can be broadly classified into ``Basic Image Manipulations'' and ``Deep Learning Approaches'', and there are also ``Meta Learning'' to explore optimal DAs metaphorically. The DAs used in HAR falls under ``Geometric Transformations'' or ``Random Erasing'' of ``Basic Image Manipulations''. In other words, only a small part of the DA methods studied in image recognition is being used in HAR.

In this study, I focus on synthetic-style DA methods, which have not been utilized in HAR in the past. Synthetic-style DA methods in image recognition are widely used: miuxp \cite{Zhang2018} and RICAP \cite{Takahashi2018}. mixup \cite{Zhang2018} is a method that combines multiple training data. Given two labeled data $(x_1, y_1)$ and $(x_2,y_2)$, mixup generates the synthetic data $(\tilde{x},\tilde{y})$ by the following formula (\ref{eq:mixup}).
\begin{eqnarray}
    \label{eq:mixup}
    \tilde{x}_{mixup} & = & \lambda x_1+(1-\lambda)x_2 \nonumber \\
    \tilde{y}_{mixup} & = & \lambda y_1+(1-\lambda)y_2
\end{eqnarray}
where $\lambda \sim Beta(\alpha, \alpha)$, for $\alpha \in (0, \infty) $, and $\lambda \in [0, 1]$. The $\alpha$ is a hyperparameter of the mixup. From the above equation, it can be said that mixup is a method of combining two inputs and outputs by weighted averaging using a random weight value $\lambda$ based on beta distribution. The important point of mixup is that not only the input $x$ but also the output $y$ are combined by weighted averaging. By this process, data that does not exist in the training data, which is the middle of the two data, is generated, and a middle label is generated.

RICAP\cite{Takahashi2018} is a method to generate a composite image by cutting out randomly determined rectangular regions from each of four training image data, and then combining them side by side. As well as mixup, the output $y$ is synthesized with a weighted average according to the size of the cut out rectangular regions.

\subsection{Advanced DA approaches}
Shorten et al. \cite{shorten2019} describes ``Deep Learning Approaches'' that use deep learning models to perform DA, such as a method for augmenting data by automatically generating data \cite{Lim2018}, and a method for augmenting data by style transformation \cite{Jackson2019}. In recent years, meta-learning methods, such as AutoAugment \cite{Cubuk2019}, which uses reinforcement learning to search for the best strategy from multiple DA methods, RandAugment \cite{Cubuk2020}, which is faster by performing this randomly, and Adversarial AutoAugment \cite{Zhang2020}, which is faster by adversarial training, have been proposed. A method called AugMix \cite{Hendrycks2020}, which improves robustness by synthesizing multiple DAs, has also been proposed.

Advanced DA approaches are methods that explore combinations of DA methods in ``Basic Image Manipulations''; therefore it is still important to develop new methods for ``Basic Image Manipulations''. In this paper, I propose a new synthetic-style DA method and an optimal ensemble method and training algorithm, this is a position of this study.

\section{Proposed Method}
\subsection{Outline}
\begin{figure}[tb]
  \centering
  \includegraphics[width=8.5cm]{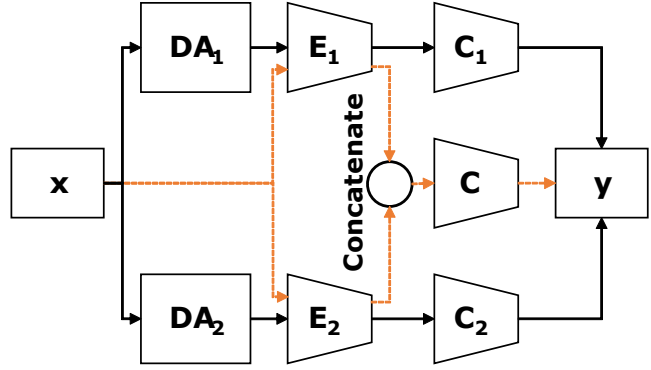}
  \caption{Model architecture and learning procedure of my proposed method. Pre-training phase ($E_1, E_2, C_1, C_2 | DA_1, DA_2$): train each network using two data augmentation policies (black solid line). Classifier-training phase ($ \hat{E}_1, \hat{E}_2, C | None$): train a new classifier C without data augmentation while freezing feature extractors $\{\hat{E}_1, \hat{E}_2\}$ (orange dashed line). Prediction phase ($E_1, E_2, C | None$): predict activity labels (orange dashed line). }
  \label{fig:ensemble}
\end{figure}
My proposed method consists of three components: Octave Mix, a new synthetic-style DA method; DA Ensemble Model which uses feature extractors trained by multiple DAs together; and DAR-FFE which pre-trains using DAs and additionally trains only the classifier part without using DAs. An overview diagram is illusrated in Figure \ref{fig:ensemble}. The individual details are described in the following sections.

\subsection{Octave Mix (OctMix)}
Octave Mix is inspired by Octave Convolution \cite{Chen2019}, which perform convolution after decomposing low and high frequency components, and applies it to synthetic-style DA. The Octave Mix algorithm for mini-batch input is shown in Algorithm. \ref{alg1}. First, a Low Pass Filter (LPF) and a High Pass Filter (HPF) are applied to the input to decompose the low frequency component LPF(x) and the high frequency component HPF(x). The LPF(x) and the HPF(x) with randomized order are combined. The resulting two composite waveforms are combined using a weighted sum with the coefficient $\lambda$ as the weight.

\begin{figure}[!t]
    \begin{algorithm}[H]
        \caption{Octave Mix algorithm for mini batch}
        \label{alg1}
        \begin{algorithmic}[1]
        \renewcommand{\algorithmicrequire}{\textbf{Input:}}
        \renewcommand{\algorithmicensure}{\textbf{Output:}}
        \REQUIRE A mini batch training set $(X, Y) = (\{x_i\}^n_{i=1}, \{y_i\}^n_{i=1})$; hyper parameter $\alpha \in (0, \infty)$; cutoff frequency $f_c$;
        \STATE $X^{low} \leftarrow LPF(X, f_c)$
        \STATE $X^{high} \leftarrow HPF(X, f_c)$
        \STATE $I \leftarrow \{i\}^{n}_{i=1}$
        \STATE Shuffle the indices $I$
        \STATE $\lambda \sim Beta(\alpha, \alpha)$
        \FOR {$i = 1$ to $n$}
          \STATE $j \leftarrow I_i$
          \STATE $g_1 \leftarrow x^{low}_i + x^{high}_j$
          \STATE $g_2 \leftarrow x^{low}_j + x^{high}_i$
          \STATE $\tilde{x}_i = \lambda g_1 + (1-\lambda) g_2$
          \STATE $\tilde{y}_i = \lambda y_i + (1-\lambda) y_j$
        \ENDFOR
        \ENSURE $\tilde{X} = \{\tilde{x}_1, \tilde{x}_2, ... \tilde{x}_i\}, \tilde{Y} = \{\tilde{y}_1, \tilde{y}_2, ... \tilde{y}_i\}$
        \end{algorithmic}
    \end{algorithm}
\end{figure}

Thus, given two labeled data $(x_1, y_1), (x_2, y_2)$, the data $(\tilde{x}_{octmix},\tilde{y}_{octmix})$ generated by this algorithm can be formulated by the following equation (\ref{eq:oct}).

\begin{eqnarray}
  \label{eq:oct}
  \tilde{x}_{octmix} & = & \lambda \{LPF(x_1) + HPF(x_2)\} \nonumber \\
   &  & + (1-\lambda) \{LPF(x_2)+HPF(x_1)\}, \nonumber \\
  \tilde{y}_{octmix} & = & \lambda y_1 + (1-\lambda)y_2  
\end{eqnarray}

The main idea of the Octave Mix is to perform frequency decomposition before synthesizing $x$, and in addition to that, synthesize $y$ as well as mixup. Fig. \ref{fig:octmix} shows an example of generating synthetic waveforms by Octave Mix using the data $x_1$="walking" and $x_2$="jogging". The waveform of $x_1$, which is the walking data, is observed to have lower amplitude and lower frequency. When LPF and HPF are applied to $x_1$, the waveform is decomposed into a smooth walking waveform and a noise-like vibration. Similarly, when focusing on the $x2$, the amplitude and frequency of the low-frequency component is slightly higher than that of $x1$, and the amplitude of the high-frequency component is generally increased. By intersecting and combining these two waveforms, a waveform that looks like the vibration of the high-frequency component of jogging is added to the low-frequency component of walking, and a waveform that looks like the vibration of the high-frequency component of walking is added to the low-frequency component of jogging is generated. Finally, the two waveforms are combined by weighted averaging to produce a waveform according to the weight coefficient $\lambda$, as shown on the right in the figure.

There are two hyperparameters in the Octave Mix: one is $\alpha$, which is used to determine the weight value $\lambda \sim Beta(\alpha, \alpha)$. Following the mixup, the parameters of the beta distribution are unified as $\alpha$. The other is the cutoff frequency $f_c$ of LPF and HPF. Especially in HAR, the main frequency observed changes depending on the types of activities to be recognized. Therefore, it is desirable to adjust $f_c$ according to the task to be applied.

Defining each synthetic waveform as $g_1, g_2$, the Octave Mix synthesis can be transformed into the following equation (\ref{eq:oct2}).
\begin{eqnarray}
    \label{eq:oct2}
    g_1 & = & LPF(x_1) + HPF(x_2), \nonumber \\
    g_2 & = & LPF(x_2) + HPF(x_1), \nonumber \\
    \tilde{x}_{octmix} & = & \lambda g_1 + (1-\lambda) g_2, \nonumber \\
    \tilde{y}_{octmix} & = & \lambda y_1 + (1-\lambda) y_2
\end{eqnarray}
Equation (\ref{eq:oct2}) can be regarded as the equation (\ref{eq:mixup}) of mixup when $g_1 = x_1, g_2=x_2$. This means that the Octave Mix process is the same as mixup when the cutoff frequency $f_c$ is made as large as possible. Therefore, Octave Mix is an extension of mixup that includes mixup as a part.

\subsection{Ensemble augmentation model architecture}
While using Octave Mix as a DA strategy, I propose an ensemble model architecture in which multiple types of feature representations are acquired by multiple types of DAs. The model architecture is illusrated in Fig. \ref{fig:ensemble}, in which the upper path is a general deep learning model. Here, $DA_1, DA_2$ are different DA strategies, $E_1, E_2$ are feature extractors, and $C_1, C_2, C$ are classifiers. Since the internal architecture of the feature extractors and classifiers are not restricted, a variety of architectures such as VGG \cite{Simonyan2014} and ResNet \cite{He2016} can be supported. For example, in the case of VGG, the convolution-pooling layer up to just before flatten is used as the feature extractor, and the remaining fully-connected layer is used as the classifier. Therefore, the proposed method is an ensemble method in which two types of feature extractors with two different DA strategies are trained separately. Since the proposed method outputs two predictions at prediction phase, the outputs of the two feature extractors are combined to output a single prediction result using a different classifier $C$ (orange dashed line). I describe the training procedure for $C$ in the next section.

The ensemble model is based on the idea that two different DA strategies contribute to the acquisition of different feature representations. Therefore, it is necessary that $DA_1, DA_2$ are different strategies. In this study, after conducting the experiments described below, I adopted the strategy of overlaying Rotation with Octave Mix on $DA_1$ and Rotation with mixup on $DA_2$.

\subsection{Data Augmentation Revisited for Fixed Feature Extractor (DAR-FFE)}
He et al. \cite{He2019} pointed out that performing powerful data augmentation such as mixup and AutoAugment, has a possibility to enphasize a gap between the original data and the augmented data. To address this issue, they proposed DA Revisited which trained the model for N epochs on the augmented data, and then additionally trained the model for M epochs on the clean data.

In this study, inspired by DA revisited, I propose DA Revisited with Fixed Feature Extractor (DAR-FFE), in which the $E_1, E_2, C_1, C_2$ in Fig. \ref{fig:ensemble} is trained using augmented data and only $C$ is trained using the original data without DA. Although it is not discussed in He et al. literature, additional M-epochs training of the feature extractor with clean data may lead to the loss of feature representations with variations acquired by DA. Therefore, in DAR-FFE, I decided to train the feature extractor $(E_1, E_2)$ using DA in the pre-training, and train only the combined classifier $C$ using clean data in the additional training using the weights-fixed feature extractor $(\hat{E}_1, \hat{E}_2)$. Based on the above, the training procedure for the model ensembling $K$ types of DAs is shown in Algorithm \ref{alg2}.

\begin{figure}[!t]
    \begin{algorithm}[H]
        \caption{Training DA K-ensemble model by DAR-FFE}
        \label{alg2}
        \begin{algorithmic}[1]
        \renewcommand{\algorithmicrequire}{\textbf{Input:}}
        \renewcommand{\algorithmicensure}{\textbf{Output:}}
        \REQUIRE Training dataset $(X, Y)$; $K$ feature extractors $\{E_1, E_2, ... , E_K\}$; $K+1$ classifiers $\{C, C_1, C_2, ... , C_K\}$; $K$ DA policies $\{DA_1, DA_2, ... , DA_K\}$; Pre-training and classifier-training epochs $N$ and $M$;
        \FOR {$k = 1$ to $K$}
          \FOR {$epoch = 1$ to $N$}
            \STATE $(\tilde{X}_k, \tilde{Y}_k) \leftarrow DA_k(X, Y)$
            \STATE Training $E_k, C_k$ on augmented training dataset $(\tilde{X}_k, \tilde{Y}_k)$
          \ENDFOR
        \ENDFOR
        \STATE Fixing the weights of all feature extractors $\{E_1, E_2, ... , E_K\}$
        \FOR {$epoch = 1$ to $M$}
          \STATE Training $C$ on original training dataset $(X, Y)$
        \ENDFOR
        \ENSURE A trained model $\{E_1, E_2, ... , E_K, C\}$
        \end{algorithmic}
    \end{algorithm}
\end{figure}

\section{Experiments}
\subsection{Experimental settings}
\subsubsection{Datasets}
I used four public datasets (HASC \cite{Kawaguchi2011}, PAMAP2 \cite{Reiss2012a}, UCI Smartphone \cite{Anguita2013} and UniMiB SHAR \cite{Daniela2017}) summarized in Table \ref{table:dataset}. All of these are benchmark datasets for HAR using sensor data. 

\begin{table*}[tb] 
  \caption{Details of dataset for evaluation. Sensor type A denotes acceleration sensor, G denotes gyroscope, M denotes magnetic sensor, where the ``?'' denotes batch size. }
  \label{table:dataset}
  \scalebox{1.0}{
	\hbox to\hsize{\hfil
	\small 
  \begin{tabular}{cc|ccc|cccc|cc} \hline \hline
    \multirow{2}{*}{Cite} &	\multirow{2}{*}{Dataset} &	\multicolumn{3}{c|}{Num. of subjects in} &	\multicolumn{4}{c|}{Sensor} &	\multicolumn{2}{c}{Shape of} \\ 
    &	 &	train &	valid &	test &	position &	type &	axes &	sampling &	input &	output \\ \hline
    \cite{Kawaguchi2011} &	HASC &	10 &	50 &	50 &	Preferred &	A &	x, y, z &	100 Hz &	(?, 256, 3) &	(?, 6) \\  
    \cite{Reiss2012a} &	PAMAP2 &	4 &	2 &	2 &	chest, wrist, ankle &	A1, A2, G, M &	x, y, z &	100 Hz &	(?, 256, 36) &	(?, 12) \\
    \cite{Anguita2013} &	UCI Smartphone &	10 &	10 &	10 &	belt / preferred &	A, G &	x, y, z &	50 Hz &	(?, 128, 6) &	(?, 6) \\ 
    \cite{Daniela2017} &	UniMiB SHAR &	10 &	10 &	10 &	pocket &	A &	x, y, z &	50 Hz &	(?, 151, 3) &	(?, 17) \\ \hline
  \end{tabular}\hfil}
  }
\end{table*} 

HASC \cite{Kawaguchi2011} is a benchmark dataset for basic HAR using smartphone sensors. It consists of accelerometer and gyroscope measurements labeled with six basic activities (staying, walking, jogging, skipping, going up stairs, and going down stairs). I extracted data with a sampling frequency of 100 Hz from BasicActivity of the corpus from 2011 to 2013, and used only the raw data of the accelerometer. As a preprocessing step, I removed 5 seconds before and after from each measurement file, and divided the data into time series with 256 samples of frame size and 256 samples of stride. As a result, the shape of the input data is as follows: winsize=256, channels=3 (x,y,z). 5 seconds before and after the start of measurement are trimmed to remove the influence of the storage operation of the device. I did not used meta labels, such as measured divice information and personal information of subjects (e.g. gender, age, height and weight). As an experimental dataset, I used the data of 176 persons whose data could be obtained more than one frame after trimming.

PAMAP2 \cite{Reiss2012a} is a benchmark dataset in which three Wireless IMUs were worn on the chest, wrist and ankle to record daily activities. This dataset collected 8 subjects' acceleration sensor data labeled with 12 activities (other, lying, sitting, standing, walking, jogging, cycling, computer work, car driving, ascending stairs, descending stairs, ironing). I devided the data into time series with 256 samples of frame size and 256 samples of stride. From the above, the shape of the input data is as follows: winsize=256, channels = 3 IMUs * 4 types of sensors * 3 (x, y, z) = 36.

UCI Smartphone \cite{Anguita2013} is a benchmark dataset using smartphone sensors for HAR. This dataset collected 30 subjects' acceleration sensor and gyroscope data labeled with 6 activities (siting, standing, laying walking, going up stairs, going down stairs). This dataset have been divided into 128 samples for each. From the above, the shape of the input data is as follows: winsize=128, channels = 2 types of sensors * 3 (x, y, z) = 6.

UniMiB SHAR \cite{Daniela2017} is a benchmark dataset using smartphone sensors for HAR. This dataset collected 30 subjects' acceleration sensor data labeled with 17 activities (standing up from sitting, standing up from laying, walking, jogging, jumping, going up stairs, going down stairs, lying down from standing, sitting down, generic falling forward, falling rightward, generic falling backward, hitting an obstacle in the fall, falling with protection strategies, falling backward-sitting-chair, falling leftward, syncope). This dataset have been divided into 151 samples for each. From the above, the shape of the input data is as follows: winsize=151, channels = 3 (x, y, z).
\begin{figure}[tb]
  \centering
  \includegraphics[width=8.5cm]{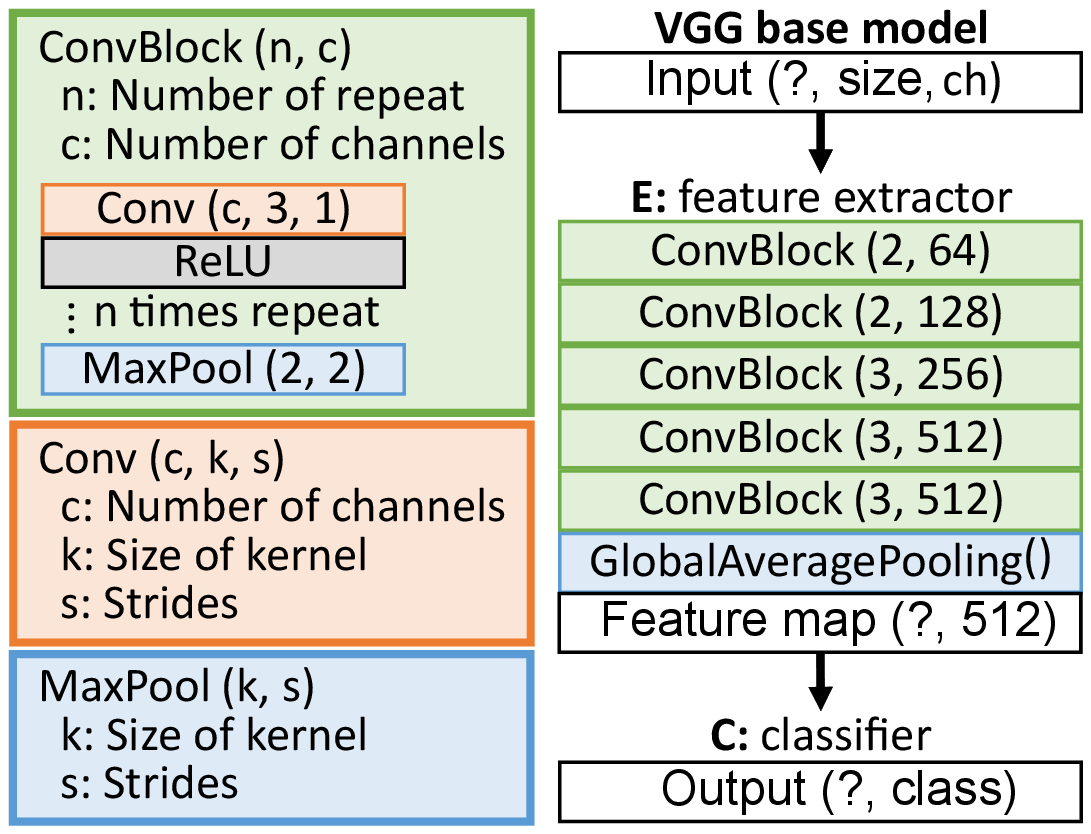}
  \caption{Model architectures for experiments. The feature extractor part $E$ including five ConvBlocks outputs (?, 512) shape feature map via a global average pooling layer, where the ``?'' denotes batch size. The classifier part $C$ is composed of only single fully-connected layer which outputs (?, class) shape soft-max values as one-hot vector. }
  \label{fig:vgg}
\end{figure}

\subsubsection{Model training}
My proposed method can be applied to any deep learning model architectures. My method used divided a general CNN model into a feature extractor and a classifier as illusrated in Fig. \ref{fig:ensemble}. In this study, I adopted the VGG architecture, which has been validated in my previous study \cite{Hasegawa2019}, and used the architecture illustrated in Fig. \ref{fig:vgg}. My previous study adopted the original VGG architecture \cite{Simonyan2014}, but in this study, to reduce the effect of $C$, I changed the flatten to GlobalAveragePooling and changed the fully-connected layer of $C$ to only single layer.

The model is trained on Adam \cite{Kingma2015} for 300 epochs with a learning rate $\eta=0.001$. I have confirmed that the training converges in 300 epochs in all conditions. In the case of using DA, I applied DA to the input data $(X, Y)$ with a probability of 50\% to obtain $(\tilde{X}, \tilde{Y})$, and then combined $(X, Y)$ and $(\tilde{X}, \tilde{Y})$ as input.

\subsubsection{Metrics}
As Gholamiangonabadi el al. \cite{Gholamiangonabadi2020} pointed out, to evaluate sensor-based HAR, dataset should be divided by subjects. Assuming that labeled sensor data of prediction-target user could not be obtained in the real use case, I evaluated by subject-base-hold-out validation which divided dataset into training, validation, testing by subjects. The number of subjects included in each datasets is shown in Table \ref{table:dataset}. Subjects are selected by random sampling. (1) Sampling of subjects, (2) dividing of the dataset, and (3) training and accuracy evaluation of each method are considered as one trial. The estimation accuracy between methods is compared by discussing the average of the results of 10 trials. Since there is a bias in the number of data between labels in some datasets, I use the average f-score as an evaluation index in addition to accuracy.

\subsection{Parameter tuning of DA}
The purpose of this study is to verify the effectiveness of synthetic-style DA methods and to develop new methods. I compared my proposed method with typical synthetic-style DA methods: mixup \cite{Zhang2018} and RICAP\cite{Takahashi2018}. As a DA method based on simple geometric transformations, I applied Rotation, which has been shown to be effective for HAR in the study of Um et al. \cite{Um2017}. Because of waveform data, I adopted to combine two waveforms back to forth in the time series direction as a RICAP procedure. Similarly, mixup was a weighted average of two waveforms.

\begin{table*}[tb] 
  \caption{Hyper-parameter tuning result using HASC dataset (the best accuracy of all epochs for validation set). }
  \label{table:tuning}
  \scalebox{1.}{
  \hbox to\hsize{\hfil
  \begin{tabular}{c|cc|cc|cccccc} \hline \hline
        &	\multirow{2}{*}{None} &	\multirow{2}{*}{Rotation} &	Rotation &	Rotation &	\multicolumn{6}{c}{Rotation \& OctMix} \\ 
    $\alpha$ &	 &	 &	\& mixup &	\& RICAP &	$f_c=0.1$ &	$1.1$ &	$2.1$ &	$3.1$ &	$4.1$ &	$5.1$ \\ \hline
    0.5 &	\multirow{3}{*}{71.8\%} &	\multirow{3}{*}{79.2\%} &	79.9\% &	79.8\% &	78.7\% &	81.0\% &	\ub{81.3\%} &	80.6\% &	80.8\% &	80.8\% \\ 
    1.0 &	 &	 &	79.7\% &	79.8\% &	78.6\% &	80.9\% &	81.2\% &	80.9\% &	80.7\% &	80.6\% \\ 
    5.0 &	 &	 &	\ub{80.2\%} &	\ub{80.0\%} &	79.7\% &	81.2\% &	80.8\% &	80.6\% &	80.2\% &	80.4\% \\ \hline
  \end{tabular}\hfil}
  }
\end{table*} 
\begin{table*}[tb] 
  \caption{Comparison results of estimation accuracy for four dataset (accuracy and average f-score for test set [\%]). }
  \label{table:compare_dataset}
  \scalebox{1.0}{
	\hbox to\hsize{\hfil
	\small 
  \begin{tabular}{c|cc|cc|cc|cc} \hline \hline
    \multirow{2}{*}{DA} &	\multicolumn{2}{c|}{HASC} &	\multicolumn{2}{c|}{PAMAP} &	\multicolumn{2}{c|}{UCI Smartphone} &	\multicolumn{2}{c}{UniMiB SHAR} \\ 
    &	Accuracy &	F-score &	Accuracy &	F-score &	Accuracy &	F-score &	Accuracy &	F-score \\ \hline
   None &	68.1($\pm$4.0) &	68.1($\pm$4.1) &	52.8($\pm$12.1) &	48.2($\pm$11.5) &	80.3($\pm$13.5) &	78.7($\pm$17.4) &	60.3($\pm$3.3) &	51.1($\pm$3.5) \\ 
   Rotation(Rot.) &	76.6($\pm$ 1.9) &	76.9($\pm$ 1.9) &	61.8($\pm$11.4) &	59.8($\pm$10.2) &	88.2($\pm$4.6) &	88.3($\pm$4.9) &	68.1($\pm$2.9) &	57.5($\pm$2.6) \\ 
   Rot.\&mixup &	77.7($\pm$2.9) &	77.6($\pm$3.0) &	\ux{66.1($\pm$9.8)} &	\ux{64.6($\pm$10.4)} &	89.7($\pm$3.2) &	90.1($\pm$2.9) &	\ux{72.0($\pm$2.6)} &	\ux{61.9($\pm$2.6)} \\ 
   Rot.\&RICAP &	77.8($\pm$2.1) &	77.9($\pm$2.0) &	63.3($\pm$10.1) &	60.7($\pm$9.8) &	\ux{90.0($\pm$3.1)} &	\ux{90.5($\pm$2.9)} &	69.2($\pm$2.0) &	58.4($\pm$2.9) \\ 
   Rot.\&OctMix &	\ux{79.3($\pm$1.4)} &	\ux{79.5($\pm$1.4)} &	62.0($\pm$11.0) &	59.0($\pm$10.8) &	89.7($\pm$2.4) &	90.1($\pm$2.3) &	67.9($\pm$2.6) &	58.7($\pm$2.3) \\ 
   Mine &	\ub{80.9($\pm$1.5)} &	\ub{81.0($\pm$1.5)} &	\ub{66.5($\pm$10.6)} &	\ub{64.7($\pm$10.5)} &	\ub{90.8($\pm$3.2)} &	\ub{91.2($\pm$3.0)} &	\ub{75.1($\pm$1.9)} &	\ub{65.4($\pm$2.4)} \\ \hline
  \end{tabular}\hfil}}
\end{table*} 

For subsequent comparisons, the HASC dataset was first used to tune the hyperparameters of each DA method: $\alpha$, the parameter of the beta distribution used to determine the weights during synthesis, was used for mixup and RICAP, while $\alpha$ plus the cutoff frequency $f_c$ were the hyperparameters to be tuned for Octave Mix. The evaluation of the tuning was based on the accuracy of the vaidation data.

Table \ref{table:tuning} shows the results of hyperparameter tuning. Upper five methods trained the simple model
without DA (None) and with rotation DA (Rotation), mixup after rotation (Rot.\& mixup), RICAP after rotation (Rot.\& RICAP), and Octave Mix after rotation (Rot.\& OctMix). Mine was combination of my proposed three components (Octave Mix, ensemble model, and DAR-FFE). First, Rotation improved the estimation accuracy by 7.4\% from None.  Mixup and RICAP improved the estimation accuracy by 1.0\% for $\alpha=5.0$. Furthermore, Octave Mix improved the estimation accuracy by 2.0\% from Rotation with $\alpha=0.5, f_c=2.1$. Thus, the effectiveness of the synthetic-style DA methods mixup and RICAP was revealed, and my proposed Octave Mix further improved the accuracy. As for the hyperparameters, the effect of $\alpha$ was not so large, and $f_c$ did not make much difference if it was greater than 1. In subsequent experiments, I will discuss mixup and RICAP with $\alpha$ set to 5.0, and Octave Mix with $\alpha=0.5, f_c=2.1$.

\subsection{Evaluation using four datasets}
Table \ref{table:compare_dataset} shows the results of the evaluation of my proposed method on four different benchmark datasets. The results were evaluated on the test data using the hyperparameters determined in the previous section. The best results were marked with an underscore and boldface, and the second best results were marked with an underscore.

My proposed method achieved the highest average f-score, and 4.1\% for HASC, 4.9\% for PAMAP, 2.9\% for UCI Smartphone, and 7.9\% for UniMiB SHAR improved compared to Rotation alone. Although which synthetic-style DA was effective was different for each target task, it was found that ensembling with Octave Mix could improve the accuracy.

\subsection{Ablation study}
Table \ref{table:ablation_study} shows the results of the ablation study using the HASC dataset. The best accuracy and average f-score were obtained for my proposed method (9). Comparing effectiveness for each proposed component using (2) to (5), the effect of introducing Octave Mix was particularly significant, improving the f-score by 2.6\% compared to Rotation alone (2). DAR-FFE also had a slight effect, improving the average f-score by 0.7\%. On the other hand, (4) and (6), which were simple ensembles, showed a tendency to decrease the average f-score. Therefore, it was found that simple ensembling of DAs does not lead to the acquisition of feature representations with variation. Note that simple ensembling was a method which trained $(E_1, E_2, C)$ all together using two DA policies to evaluate the effect of un-using DAR-FFE.

In (6) to (9), I consider the effect of missing one component. In (7), the model was trained using single DA (Rot.\&OctMix) without ensembling the model, and then only the classifier was retrained on the original data. Despite the fact that the number of model parameters was the same as the simple methods (1)-(3), this method improved the accuracy by 3.2\% in f-score from Rotation alone (2) and by 0.6\% from Rot.\&OctMix (3). This was equivalent to the accuracy of (8), where only the classifier was additionally trained after ensembling RICAP and mixup. The proposed method (9) further improved the accuracy by about 1\%.
\begin{table*}[tb] 
    \caption{Results of ablation study using HASC dataset (accuracy and average f-score for test set [\%]). }
    \label{table:ablation_study}
    \scalebox{1.0}{
		\hbox to\hsize{\hfil
		\small 
    \begin{tabular}{c|cccc|cc|c} \hline \hline
      & \multicolumn{4}{c|}{Methods}   &  \multirow{2}{*}{Accuracy} &  \multirow{2}{*}{F-score} & \multirow{2}{*}{Note} \\
       &	Rotation &	OctMix &	Ensemble &	DAR-FFE &	 &	 & \\ \hline
       (1) &	 &	 &	 &	 &	68.1($\pm$4.0) &	68.1($\pm$4.1) &	Without DA \\ 
       (2) &	\checkmark &	 &	 &	 &	76.6($\pm$1.9) &	76.9($\pm$1.9) &	With Rotation \\ 
       (3) &	\checkmark &	\checkmark &	 &	 &	79.3($\pm$1.4) &	79.5($\pm$1.4) &	With Rot. \& OctMix \\ 
       (4) &	\checkmark &	 &	\checkmark &	 &	76.1($\pm$2.6) &	76.2($\pm$2.7) &	Ensemble Rot.\&RICAP and Rot.\&mixup \\ 
       (5) &	\checkmark &	 &	 &	\checkmark &	77.3($\pm$1.9) &	77.6($\pm$1.9) &	DAR-FFE from (2) \\ 
       (6) &	\checkmark &	\checkmark &	\checkmark &	 &	74.7($\pm$2.8) &	74.8($\pm$3.3) &	Ensemble Rot.\&OctMix and Rot.\&mixup \\ 
       (7) &	\checkmark &	\checkmark &	 &	\checkmark &	79.9($\pm$1.9) &	80.1($\pm$1.9) &	DAR-FFE from (3) \\ 
       (8) &	\checkmark &	 &	\checkmark &	\checkmark &	\ux{80.0($\pm$2.0)} &	\ux{80.2($\pm$2.0)} &	DAR-FFE from (4) \\ 
       (9) &	\checkmark &	\checkmark &	\checkmark &	\checkmark &	\ub{80.9($\pm$1.5)} &	\ub{81.0($\pm$1.5)} &	DAR-FFE from (6) \\ \hline
    \end{tabular}\hfil}}
\end{table*}

\subsection{Effects of the amount of training data}
\begin{figure}[tb]
    \centering
    \includegraphics[width=8.5cm]{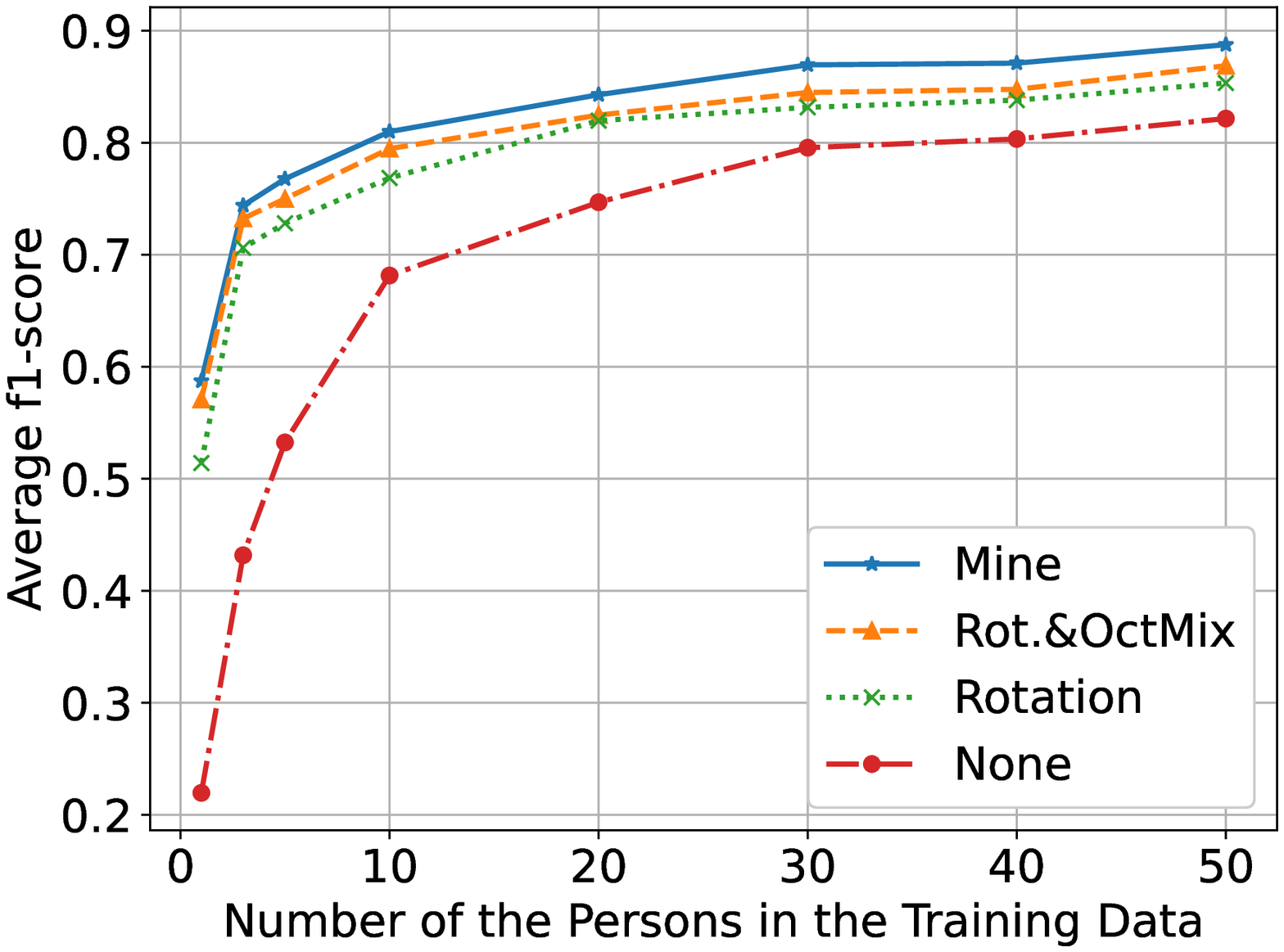}
    \caption{Effect of changing the number of subjects in the training set (average f-score for test set [\%]). }
    \label{fig:numofusers}
\end{figure}

Fig. \ref{fig:numofusers} shows the change in the average f-score when the number of subjects in the training data is changed. The proposed method works effectively regardless of the amount of training data. In addition, the difference between Mine and Rot.\&OctMix was relatively small when the number of subjects is small (less than 10 persons), and the difference became more pronounced as the number of subjects increases. In other words, the effect of Octave Mix was large when the amount of data is small, and the effect of ensemble became larger as the number of subjects increases.

\subsection{Comparison K-ensemble}
Table \ref{table:kensemble} shows the experimental results of my proposed method with different DA combinations. My proposed method was an ensemble method of Octave Mix and mixup (b) in the table. The estimation accuracy of the proposed method is higher than that of the other combinations (a) and (c). In addition, it had a possibility to increase the number of DA combinations would improve the accuracy. I evaluated the method combining three DA policies (d), but there was almost no change in accuracy from (b). Therefore, combining two DA policieswere sufficient for the DA variation adopted in this study. In the future, combining any other DA methods to acquire a variety of feature representation, may be able to improve the accuracy by increasing the number of combinations.

\begin{table}[tb] 
    \caption{Effect of changing ensemble patterns (accuracy and average f-score for test set [\%]). }
    \label{table:kensemble}
    \scalebox{1.0}{
    \hbox to\hsize{\hfil
    \begin{tabular}{c|ccc|cc} \hline \hline
      &	\multicolumn{3}{c|}{Rotation \&} &	\multirow{2}{*}{Accuracy} &	\multirow{2}{*}{F-score} \\ 
      &	mixup &	RICAP &	OctMix &	 &	 \\ \hline
      (a) &	\checkmark &	\checkmark &	 &	80.0($\pm$2.0) &	80.2($\pm$2.0) \\
      (b) &	\checkmark &	 &	\checkmark &	\ux{80.9($\pm$1.5)} &	\ux{81.0($\pm$1.5)} \\
      (c) &	 &	\checkmark &	\checkmark &	80.5($\pm$1.7) &	80.8($\pm$1.6) \\
      (d) &	\checkmark &	\checkmark &	\checkmark &	\ub{80.9($\pm$1.7)} &	\ub{81.1($\pm$1.6)} \\ \hline    
    \end{tabular}\hfil}
    }
\end{table}

\section{Conclusion}
In this study, I proposed and validated a new DA method Octave Mix for sensor-based HAR, a model architecture for ensembling DAs, and a method for additional training on the original data (DAR-FFE). My proposed method is a DA method that combines multiple input data and improves the conventional mixup method by using frequency decomposition. As a result of experiments, I confirmed that the three components of my proposed method (Octave Mix, ensemble model, and DAR-FFE) could improve the estimation accuracy of HAR. In the future, I expect that the three components of my proposed method will be applied to various problems and problem settings in different fields.

\bibliographystyle{IEEEtran}
\bibliography{mybib}

%\bibliographystyle{unsrtnat}
%\bibliography{references}  %%% Uncomment this line and comment out the ``thebibliography'' section below to use the external .bib file (using bibtex) .

%%% Uncomment this section and comment out the \bibliography{references} line above to use inline references.
% \begin{thebibliography}{1}

% 	\bibitem{kour2014real}
% 	George Kour and Raid Saabne.
% 	\newblock Real-time segmentation of on-line handwritten arabic script.
% 	\newblock In {\em Frontiers in Handwriting Recognition (ICFHR), 2014 14th
% 			International Conference on}, pages 417--422. IEEE, 2014.

% 	\bibitem{kour2014fast}
% 	George Kour and Raid Saabne.
% 	\newblock Fast classification of handwritten on-line arabic characters.
% 	\newblock In {\em Soft Computing and Pattern Recognition (SoCPaR), 2014 6th
% 			International Conference of}, pages 312--318. IEEE, 2014.

% 	\bibitem{hadash2018estimate}
% 	Guy Hadash, Einat Kermany, Boaz Carmeli, Ofer Lavi, George Kour, and Alon
% 	Jacovi.
% 	\newblock Estimate and replace: A novel approach to integrating deep neural
% 	networks with existing applications.
% 	\newblock {\em arXiv preprint arXiv:1804.09028}, 2018.

% \end{thebibliography}

\end{document}